\documentclass[10pt,twocolumn,letterpaper]{article}

\usepackage{cvpr}              %
\usepackage{graphicx}

\newcommand{\model}{ART\xspace}

\newcommand{\myparagraph}[1]{\vspace{0.1cm}\noindent\textbf{#1}}
\renewcommand{\paragraph}[1]{\vspace{0.1cm}\noindent\textbf{#1}}

\definecolor{MyDarkBlue}{rgb}{0,0.08,1}
\definecolor{MyAqua}{rgb}{0,0.7,0.7}
\definecolor{MyDarkGreen}{rgb}{0.02,0.6,0.02}
\definecolor{MyDarkRed}{rgb}{0.8,0.02,0.02}
\definecolor{MyDarkOrange}{rgb}{0.40,0.2,0.02}
\definecolor{MyPurple}{RGB}{111,0,255}
\definecolor{MyRed}{rgb}{1.0,0.0,0.0}
\definecolor{MyGold}{rgb}{0.75,0.6,0.12}
\definecolor{MyDarkgray}{rgb}{0.66, 0.66, 0.66}

\definecolor{Cardinal}{rgb}{0.549,0.082,0.082}
\usepackage[accsupp]{axessibility}  %

\definecolor{cvprblue}{rgb}{0.21,0.49,0.74}
\usepackage[pagebackref,breaklinks,colorlinks,allcolors=cvprblue]{hyperref}

\def\paperID{}
\def\confName{CVPR}
\def\confYear{2026}

\makeatletter
\newcommand{\printfnsymbol}[1]{%
    \textsuperscript{\@fnsymbol{#1}}%
}
\makeatother

\newcommand\rurl[1]{%
  \href{https://#1}{\nolinkurl{#1}}%
}

\title{\model: Articulated Reconstruction Transformer}

\author{
Zizhang Li\textsuperscript{1,2}\printfnsymbol{1} \; 
Cheng Zhang\textsuperscript{1} \;
Zhengqin Li\textsuperscript{1} \;
Henry Howard-Jenkins\textsuperscript{1} \;
Zhaoyang Lv\textsuperscript{1} \;
Chen Geng\textsuperscript{2} \;
\\
Jiajun Wu\textsuperscript{2} \;
Richard Newcombe\textsuperscript{1} \;
Jakob Engel\textsuperscript{1} \;
Zhao Dong\textsuperscript{1} \;
\\[0.5em]
\textsuperscript{1}Reality Labs Research, Meta \quad
\textsuperscript{2}Stanford University
\\
\\
{\url{https://kyleleey.github.io/ART/}}
}

\begin{document}

\newcommand{\ba}{\boldsymbol{a}}\newcommand{\bA}{\boldsymbol{A}}
\newcommand{\bb}{\boldsymbol{b}}\newcommand{\bB}{\boldsymbol{B}}
\newcommand{\bc}{\boldsymbol{c}}\newcommand{\bC}{\boldsymbol{C}}
\newcommand{\bd}{\boldsymbol{d}}\newcommand{\bD}{\boldsymbol{D}}
\newcommand{\be}{\boldsymbol{e}}\newcommand{\bE}{\boldsymbol{E}}
\newcommand{\bff}{\boldsymbol{f}}\newcommand{\bF}{\boldsymbol{F}} %
\newcommand{\bg}{\boldsymbol{g}}\newcommand{\bG}{\boldsymbol{G}}
\newcommand{\bh}{\boldsymbol{h}}\newcommand{\bH}{\boldsymbol{H}}
\newcommand{\bi}{\boldsymbol{i}}\newcommand{\bI}{\boldsymbol{I}}
\newcommand{\bj}{\boldsymbol{j}}\newcommand{\bJ}{\boldsymbol{J}}
\newcommand{\bk}{\boldsymbol{k}}\newcommand{\bK}{\boldsymbol{K}}
\newcommand{\bl}{\boldsymbol{l}}\newcommand{\bL}{\boldsymbol{L}}
\newcommand{\bm}{\boldsymbol{m}}\newcommand{\bM}{\boldsymbol{M}}
\newcommand{\bn}{\boldsymbol{n}}\newcommand{\bN}{\boldsymbol{N}}
\newcommand{\bo}{\boldsymbol{o}}\newcommand{\bO}{\boldsymbol{O}}
\newcommand{\bp}{\boldsymbol{p}}\newcommand{\bP}{\boldsymbol{P}}
\newcommand{\bq}{\boldsymbol{q}}\newcommand{\bQ}{\boldsymbol{Q}}
\newcommand{\br}{\boldsymbol{r}}\newcommand{\bR}{\boldsymbol{R}}
\newcommand{\bs}{\boldsymbol{s}}\newcommand{\bS}{\boldsymbol{S}}
\newcommand{\bt}{\boldsymbol{t}}\newcommand{\bT}{\boldsymbol{T}}
\newcommand{\bu}{\boldsymbol{u}}\newcommand{\bU}{\boldsymbol{U}}
\newcommand{\bv}{\boldsymbol{v}}\newcommand{\bV}{\boldsymbol{V}}
\newcommand{\bw}{\boldsymbol{w}}\newcommand{\bW}{\boldsymbol{W}}
\newcommand{\bx}{\boldsymbol{x}}\newcommand{\bX}{\boldsymbol{X}}
\newcommand{\by}{\boldsymbol{y}}\newcommand{\bY}{\boldsymbol{Y}}
\newcommand{\bz}{\boldsymbol{z}}\newcommand{\bZ}{\boldsymbol{Z}}

\newcommand{\balpha}{\boldsymbol{\alpha}}\newcommand{\bAlpha}{\boldsymbol{\Alpha}}
\newcommand{\bbeta}{\boldsymbol{\beta}}\newcommand{\bBeta}{\boldsymbol{\Beta}}
\newcommand{\bgamma}{\boldsymbol{\gamma}}\newcommand{\bGamma}{\boldsymbol{\Gamma}}
\newcommand{\bdelta}{\boldsymbol{\delta}}\newcommand{\bDelta}{\boldsymbol{\Delta}}
\newcommand{\bepsilon}{\boldsymbol{\epsilon}}\newcommand{\bEpsilon}{\boldsymbol{\Epsilon}}
\newcommand{\bzeta}{\boldsymbol{\zeta}}\newcommand{\bZeta}{\boldsymbol{\Zeta}}
\newcommand{\beeta}{\boldsymbol{\eta}}\newcommand{\bEta}{\boldsymbol{\Eta}} %
\newcommand{\btheta}{\boldsymbol{\theta}}\newcommand{\bTheta}{\boldsymbol{\Theta}}
\newcommand{\biota}{\boldsymbol{\iota}}\newcommand{\bIota}{\boldsymbol{\Iota}}
\newcommand{\bkappa}{\boldsymbol{\kappa}}\newcommand{\bKappa}{\boldsymbol{\Kappa}}
\newcommand{\blambda}{\boldsymbol{\lambda}}\newcommand{\bLambda}{\boldsymbol{\Lambda}}
\newcommand{\bmu}{\boldsymbol{\mu}}\newcommand{\bMu}{\boldsymbol{\Mu}}
\newcommand{\bnu}{\boldsymbol{\nu}}\newcommand{\bNu}{\boldsymbol{\Nu}}
\newcommand{\bxi}{\boldsymbol{\xi}}\newcommand{\bXi}{\boldsymbol{\Xi}}
\newcommand{\bomikron}{\boldsymbol{\omikron}}\newcommand{\bOmikron}{\boldsymbol{\Omikron}}
\newcommand{\bpi}{\boldsymbol{\pi}}\newcommand{\bPi}{\boldsymbol{\Pi}}
\newcommand{\brho}{\boldsymbol{\rho}}\newcommand{\bRho}{\boldsymbol{\Rho}}
\newcommand{\bsigma}{\boldsymbol{\sigma}}\newcommand{\bSigma}{\boldsymbol{\Sigma}}
\newcommand{\btau}{\boldsymbol{\tau}}\newcommand{\bTau}{\boldsymbol{\Tau}}
\newcommand{\bypsilon}{\boldsymbol{\ypsilon}}\newcommand{\bYpsilon}{\boldsymbol{\Ypsilon}}
\newcommand{\bphi}{\boldsymbol{\phi}}\newcommand{\bPhi}{\boldsymbol{\Phi}}
\newcommand{\bchi}{\boldsymbol{\chi}}\newcommand{\bChi}{\boldsymbol{\Chi}}
\newcommand{\bpsi}{\boldsymbol{\psi}}\newcommand{\bPsi}{\boldsymbol{\Psi}}
\newcommand{\bomega}{\boldsymbol{\omega}}\newcommand{\bOmega}{\boldsymbol{\Omega}}

\newcommand{\nA}{\mathbb{A}}
\newcommand{\nB}{\mathbb{B}}
\newcommand{\nC}{\mathbb{C}}
\newcommand{\nD}{\mathbb{D}}
\newcommand{\nE}{\mathbb{E}}
\newcommand{\nF}{\mathbb{F}}
\newcommand{\nG}{\mathbb{G}}
\newcommand{\nH}{\mathbb{H}}
\newcommand{\nI}{\mathbb{I}}
\newcommand{\nJ}{\mathbb{J}}
\newcommand{\nK}{\mathbb{K}}
\newcommand{\nL}{\mathbb{L}}
\newcommand{\nM}{\mathbb{M}}
\newcommand{\nN}{\mathbb{N}}
\newcommand{\nO}{\mathbb{O}}
\newcommand{\nP}{\mathbb{P}}
\newcommand{\nQ}{\mathbb{Q}}
\newcommand{\nR}{\mathbb{R}}
\newcommand{\nS}{\mathbb{S}}
\newcommand{\nT}{\mathbb{T}}
\newcommand{\nU}{\mathbb{U}}
\newcommand{\nV}{\mathbb{V}}
\newcommand{\nW}{\mathbb{W}}
\newcommand{\nX}{\mathbb{X}}
\newcommand{\nY}{\mathbb{Y}}
\newcommand{\nZ}{\mathbb{Z}}

\newcommand{\cA}{\mathcal{A}}
\newcommand{\cB}{\mathcal{B}}
\newcommand{\cC}{\mathcal{C}}
\newcommand{\cD}{\mathcal{D}}
\newcommand{\cE}{\mathcal{E}}
\newcommand{\cF}{\mathcal{F}}
\newcommand{\cG}{\mathcal{G}}
\newcommand{\cH}{\mathcal{H}}
\newcommand{\cI}{\mathcal{I}}
\newcommand{\cJ}{\mathcal{J}}
\newcommand{\cK}{\mathcal{K}}
\newcommand{\cL}{\mathcal{L}}
\newcommand{\cM}{\mathcal{M}}
\newcommand{\cN}{\mathcal{N}}
\newcommand{\cO}{\mathcal{O}}
\newcommand{\cP}{\mathcal{P}}
\newcommand{\cQ}{\mathcal{Q}}
\newcommand{\cR}{\mathcal{R}}
\newcommand{\cS}{\mathcal{S}}
\newcommand{\cT}{\mathcal{T}}
\newcommand{\cU}{\mathcal{U}}
\newcommand{\cV}{\mathcal{V}}
\newcommand{\cW}{\mathcal{W}}
\newcommand{\cX}{\mathcal{X}}
\newcommand{\cY}{\mathcal{Y}}
\newcommand{\cZ}{\mathcal{Z}}

\newcommand{\figref}[1]{Fig.~\ref{#1}}
\newcommand{\secref}[1]{Section~\ref{#1}}
\newcommand{\algref}[1]{Algorithm~\ref{#1}}
\newcommand{\eqnref}[1]{Eq.~\eqref{#1}}
\newcommand{\tabref}[1]{Table~\ref{#1}}

\def\mc{\mathcal}
\def\mb{\boldsymbol}

\newcommand{\T}{^{\raisemath{-1pt}{\mathsf{T}}}}

\newcommand{\Perp}{\perp\!\!\! \perp}

\makeatletter
\DeclareRobustCommand\onedot{\futurelet\@let@token\@onedot}
\def\@onedot{\ifx\@let@token.\else.\null\fi\xspace}
\def\eg{e.g\onedot,\xspace} \def\Eg{E.g\onedot,\xspace}
\def\ie{i.e\onedot,\xspace} \def\Ie{I.e\onedot,\xspace}
\def\cf{cf\onedot} \def\Cf{Cf\onedot}
\def\etc{etc\onedot}
\def\vs{vs\onedot}
\def\wrt{wrt\onedot}
\def\dof{d.o.f\onedot}
\def\etal{et~al\onedot}
\def\iid{i.i.d\onedot}
\def\evs{\emph{vs}\onedot}
\makeatother

\newcommand*\rot{\rotatebox{90}}

\newcommand{\boldparagraph}[1]{\vspace{0.4em}\noindent{\bf #1:}}

\definecolor{darkgreen}{rgb}{0,0.7,0}
\definecolor{lightred}{rgb}{1.,0.5,0.5}

\crefname{section}{Sec.}{Secs.}
\Crefname{section}{Section}{Sections}
\Crefname{table}{Table}{Tables}
\crefname{table}{Tab.}{Tabs.}

\renewcommand{\paragraph}{%
  \@startsection{paragraph}{4}%
  {\z@}{-0.5em}{-0.5em}%
  {\normalfont\normalsize\bfseries}%
}

\newcommand{\ignore}[1]{}{}

\twocolumn[\maketitle\begin{center}
    \includegraphics[trim={15px 5px 0px 0px}, clip, width=\linewidth]{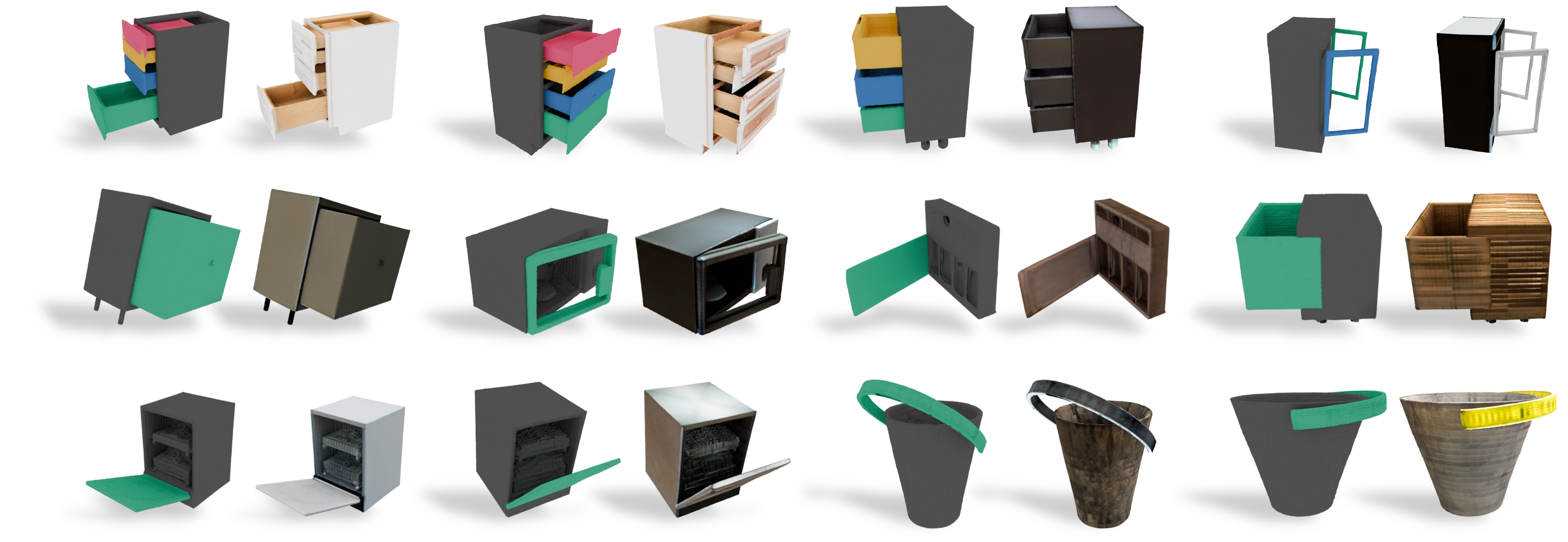}
\end{center}
\captionof{figure}{
We propose \model, a feed-forward model that reconstructs articulated objects from images by decomposing them into rigid parts. \model predicts each part's geometry, texture, and articulation structure. Here we showcase its results on diverse categories, with each image pair displaying the part-based decomposition~(left, each movable part in a unique color) and the final textured mesh~(right).
}%
\vspace{10pt}
\label{fig:teaser}\bigbreak]

\def\thefootnote{*}\footnotetext[1]{Work done during internship at Meta}\def\thefootnote{\arabic{footnote}}

\begin{abstract}

We introduce \model, \textbf{A}rticulated \textbf{R}econstruction \textbf{T}ransformer---a category-agnostic, feed-forward model that reconstructs complete 3D articulated objects from only sparse, multi-state RGB images. Previous methods for articulated object reconstruction either rely on slow optimization with fragile cross-state correspondences or use feed-forward models limited to specific object categories. In contrast, \model treats articulated objects as assemblies of rigid parts, formulating reconstruction as part-based prediction. Our newly designed transformer architecture maps sparse image inputs to a set of learnable part slots, from which \model jointly decodes unified representations for individual parts, including their 3D geometry, texture, and explicit articulation parameters. The resulting reconstructions are physically interpretable and readily exportable for simulation. Trained on a large-scale, diverse dataset with per-part supervision, and evaluated across diverse benchmarks, \model achieves significant improvements over existing baselines and establishes a new state of the art for articulated object reconstruction from image inputs.

\end{abstract}    
\section{Introduction}
\label{s:intro}

Articulated objects are ubiquitous in daily lives and central to human–scene interactions~\cite{liu2025survey}. Accurately constructing their digital replicas is important for VR/AR, robotics, and embodied AI~\cite{shen2021igibson,li2021igibson,yang2023reconstructing,deng2024articulate,li2024behavior,kim2025parahome,deng2025anymate}. While recent 3D generation and reconstruction methods have significantly advanced the automatic creation of static assets~\cite{poole2022dreamfusion,liu2023threestudio,hong2023lrm,xu2024grm,wei2024meshlrm,siddiqui2024meta,li2025lirm}, articulated objects remain challenging as they require recovering both geometry and underlying kinematic structure. Today, building such models still demands extensive expert effort, making the process labor-intensive and hard to scale, and ultimately limiting the accessibility and realism of articulated content for large-scale interactive environments.

To automate this process, we tackle the challenging problem of \textit{image-based articulated object reconstruction}: recovering a complete 3D representation for an articulated object, including geometry, texture, and its underlying articulation structure. We specifically focus on a practical, yet difficult setting: reconstructing articulated objects in a feed-forward manner from only a sparse set of multi-state RGB images. This setup is important for scalability, as dense multi-view, multi-state capture is often infeasible in real-world scenarios. However, the sparsity of inputs poses a significant challenge, requiring the inference of complex 3D shape, material, and articulation structure from limited visual cues, a task where existing articulated object reconstruction methods typically fall short.

Existing approaches to image-based articulated object reconstruction can be categorized into \textit{per-object optimization} and \textit{feed-forward learning}, both of which are ill-suited to our target setting. Per-object optimization methods~\cite{jiang2022ditto,liu2023paris,heppert2023carto,weng2024neural,liu2025building} achieve high-fidelity reconstructions but are impractically slow due to the lengthy test-time optimization; they also depend on dense observations~(often $\sim\!100$ views) and fragile cross-state matching, making them unsuitable for sparse inputs. In contrast, while feed-forward models~\cite{jiang2022opd,chen2024urdformer,liu2024singapo,gao2025meshart} offer fast inference, they are typically trained on limited datasets~(e.g., PartNet-Mobility~\cite{xiang2020sapien}, restricting them to a few categories and limiting generalization to diverse, unseen objects.

To address this gap, we introduce the Articulated Reconstruction Transformer~(\model), a category-agnostic, feed-forward model that reconstructs complete articulated 3D objects from sparse, multi-state RGB images. Our key insight is that articulated objects can be effectively represented as a collection of rigid parts, with articulation defining their kinematic relationships. Accordingly, \model formulates the reconstruction of articulated objects as a part-based prediction task.

Inspired by the success of large-scale static reconstruction models~\cite{hong2023lrm},
\model adopts a transformer architecture that maps sparse image inputs to a set of learnable part slots, each trained to capture one object part. From each slot, \model jointly decodes a unified part representation—3D geometry, texture, and explicit articulation parameters (e.g., motion type, axis, and pivot). Training with per-part supervision on a large-scale, diverse dataset yields a transferable prior, allowing \model to function as a single, unified model across categories. The part-based output is physically interpretable and directly exportable to standard simulation formats (e.g., URDF), producing simulation-ready assets.

Through comprehensive experiments, we show that \model significantly outperforms both optimization-based and feed-forward baselines. Our method sets a new state-of-the-art for articulated object reconstruction from sparse image inputs, demonstrating the potential of large-scale, part-based feed-forward models for this challenging task.

In summary, our main contributions are as follows:
\begin{itemize}
\item We tackle image-based articulated object reconstruction from sparse-view, multi-state inputs by formulating it as a part-level prediction of geometry, texture, and articulation properties.
\item We propose \model, a category-agnostic feed-forward transformer trained on large-scale articulated object datasets, capable of inferring not only per-part geometry/texture, but also kinematically consistent articulation structures.
\item We demonstrate that \model significantly outperforms both optimization-based and feed-forward baselines, establishing a new state-of-the-art for holistic articulated object reconstruction.
\end{itemize}

\section{Related Work}
\label{s:related}

\begin{figure*}
    \centering
    \includegraphics[width=\linewidth]{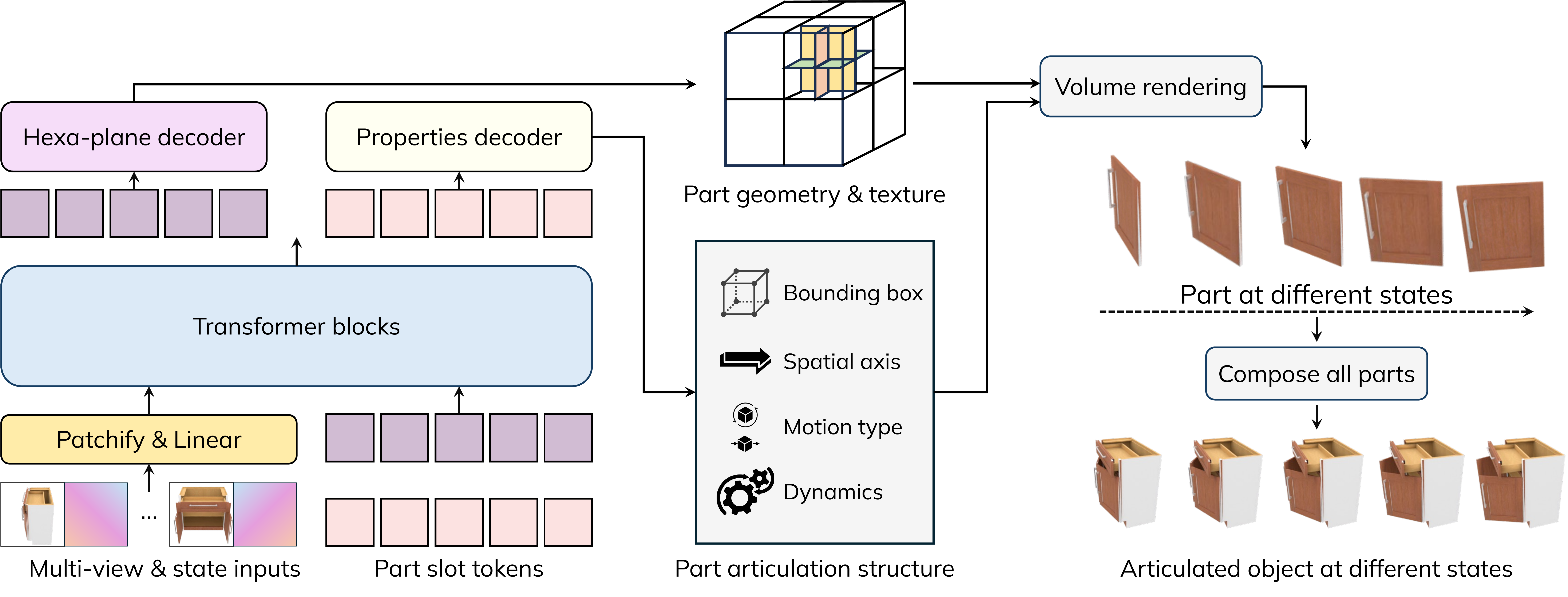}
    \caption{\textbf{Model Architecture} of \model. Multi-view, multi-state image inputs with known camera poses are tokenized and processed by a transformer alongside learnable part slot tokens. Two separate decoders then predict each part's geometry/texture and the articulation structure, and SDF volume rendering composes these components to render and reconstruct the articulated object at different states.}
    \label{fig:method}
    \vspace{-2mm}
\end{figure*}

\subsection{Articulation structure understanding}
A large body of work studies articulation understanding across diverse input modalities, including RGB~\cite{wang2024active} and RGB-D~\cite{liu2022toward,abbatematteo2019learning,jain2022distributional} images, point clouds~\cite{fu2024capt,liu2023category,wang2019shape2motion}, videos~\cite{liu2020nothing,qian2022understanding} and 3D meshes~\cite{qiu2025articulate}. 
Within this space, one line of work identifies movable parts to reveal the potential degrees of freedom of the object~\cite{jiang2022opd,sun2024opdmulti,qian2022understanding} from the given input. Another complementary line directly estimates articulation parameters—from high-level kinematic graphs to low-level joint directions, pivots, and motion angles~\cite{jiang2022ditto,li2020category,xu2022unsupervised}—with recent advances adopting generative or diffusion-based formulations to improve robustness and accuracy~\cite{liu2024singapo,lei2023nap}.
However, most prior methods treat articulation in isolation from geometry and appearance: they recover motion structure but do not reconstruct photorealistic shape and texture suitable for simulation or rendering. In contrast, our method jointly predicts articulation structure and part-level geometry/texture, producing a unified, simulation-ready representation.

\subsection{Articulated object reconstruction}
Beyond understanding alone, another line of research targets \textit{reconstruction} of articulated objects—\ie recovering geometry, texture, and articulation structure from observations. Methods typically fall into two categories. The first is the \textit{per-object optimization} methods. Many approaches formulate reconstruction as inverse rendering with neural radiance fields~\cite{mildenhall2021nerf} or 3D gaussian splatting~\cite{kerbl20233d}. These approaches iteratively optimize geometry and appearance over multi-view, multi-state sequences while inferring articulation~\cite{liu2025building,weng2024neural,liu2023paris,deng2024articulate,swaminathan2024leia,mu2021sdf}. Although often yielding high fidelity, these pipelines require very dense viewpoints or carefully staged sequences, entail lengthy per-instance optimization, and rely on fragile cross-state correspondences, making them sensitive to occlusion and initialization.

The alternative is the \textit{feed-forward prediction} models. To improve scalability, these models infer articulation without test-time optimization. For instance, SINGAPO~\cite{liu2024singapo} predicts a kinematic graph and retrieves parts to assemble full assets, while other approaches~\cite{chen2024urdformer,dai2024automated,mandi2024real2code,heppert2023carto} reconstruct articulated objects directly from single- or multi-stage image inputs. Nonetheless, data scarcity often restricts such models to a small set of categories. More recent work~\cite{gao2025partrm} leverages strong pre-trained generative prior to expand generalization, but is designed for multi-stage image generation and is typically combined with an optimization-based refinement for final reconstruction. By contrast, our approach differs remains purely feed-forward at inference from sparse RGB inputs, directly predicting a part-based 3D geometry/texture and explicit articulation parameters, and is trained on a substantially larger and more diverse dataset.

\subsection{Feed-forward 3D reconstruction}

The availability of large-scale 3D datasets~\cite{deitke2023objaverse,deitke2023objaversexl} has enabled powerful feed-forward reconstruction systems that combine scalable transformers architectures with differentiable rendering supervision~\cite{hong2023lrm,xu2024grm,li2023instant3d,jin2024lvsm,li2025lirm}. Recent work extends this approach to part-based reconstruction/generation~\cite{chen2025autopartgen,lin2025partcrafter}, indicating that structured outputs can be produced in a single forward pass.
Building on these insights, \model exploits a transformer backbone for articulated 3D reconstruction: it decomposes an object into consistent components and predicts part-level geometry and texture along with explicit articulation parameters (motion type, axis, pivot/limits) for each dynamic part, yielding a unified representation that is both photorealistic and kinematically interpretable.

\section{Articulated Reconstruction Transformer}
\label{s:method}

\subsection{Problem Formulation}
The input is a multi-view and multi-state image set $\mathcal{I} = \{\bI_{v,t}\in\nR^{H\times W\times 3} \mid v\in [1, V], t\in [1, T]\}$, where $V$ is the number of camera views and $T$ stands for the number of articulation stages (states). During training, each view has known intrinsics and extrinsics to cast sampling rays, and the object is normalized to a bounding sphere with radius $r$.

We aim to reconstruct an articulated object as a set of \textit{parts}. Let $P$ denote the number of parts~(including a static base part). For each part $p\in [1, P]$, we predict a unified representation $\cX_p$:
\begin{equation}
    \cX_p = \{\cT_p, \cA_p\}, \quad \cA_p = (\bB_p, \bC_p, \bD_p, \bO_p, \bS_p).
\end{equation}
 where $\mathcal{T}_p$ encodes geometry/texture, and
$\mathcal{A}_p$ denotes articulation parameters. The detailed definitions of these articulation parameters are given below (omitting the part index for simplicity):

\begin{itemize}
    \item $\bB=(\bB_{\mathrm{center}}, \bB_{\mathrm{size}})\in \nR^{6}$ denotes an axis-aligned bounding box in the canonical object frame, defined by its center position and the side lengths along each axis.

    \item $\bC\in \{\mathrm{static}, \mathrm{prismatic}, \mathrm{revolute}\}$ is the motion type. The base part is static by definition; all other parts are either prismatic (translational) or revolute (rotational).

    \item $\bD\in \nS^{2}$ is the joint axis direction in the canonical object frame, represented as a unit-length 3D vector.

    \item $\bO\in \nR^{3}$ is a point on the joint axis in canonical frame. For a revolute joint, it is essentially the hinge pivot. For a prismatic joint, $\bO$ is defined and predicted but is not required in inference since the direction of the axis suffices to interpret the part motion.

    \item $\bS\in \nR^{T}$ represents the normalized motion value (``dynamics'') for each input stage: angles in radians for revolute motion, or translations in object-scale units for prismatic motion. During training, $\bS$ aligns rendered images to the observed state $t$; at inference, it can be used to control the articulated configuration.
\end{itemize}

For $\cT_p$, following  transformer-based reconstruction models~\cite{hong2023lrm,wei2024meshlrm,li2025lirm}, we represent each part’s geometry and texture with a hexa-plane~\cite{li2025lirm,cao2023hexplane} parameterization (details in Section~\ref{s:method-architecture} and Section~\ref{s:method-rendering}). Concretely, we denote:
\begin{equation}
    \cT_p = \{\mathbf{T}_{p,k}^{(+,-)} ~|~ k\in\{xy,yz,xz\}\},
    \label{eq:hexa-plane}
\end{equation}
where each predicted plane $\mathbf{T}$ stores the features that will be queried during volume rendering process.
All parts in the model are predicted in a shared canonical object frame (the \emph{rest state}). Given an articulation configuration $q$~(rotational angle or translation), part $p$ can be posed by a rigid transform $T_p(q; \bC_p, \bD_p, \bO_p)$ that (i) rotates about axis $(\bD_p,\bO_p)$ when $\bC_p=\mathrm{revolute}$, or (ii) translates along direction $\bD_p$ when $\bC_p=\mathrm{prismatic}$. The base part uses the identity transform. During training, we enforce consistency between the posed prediction and the observed stage $t$ using the predicted per-state dynamics.

We allocate $P_0$ learnable part slots in the network and predict $P{\leq}P_0$ active parts. At inference time, similar to previous methods~\cite{liu2023paris,weng2024neural,liu2025building}, we assume the part count is known. In practice, existing available VLMs~\cite{comanici2025gemini,hurst2024gpt,yang2025qwen3} can also provide accurate part-count estimates.

An important design choice is to use a canonical \textit{rest state} frame for articulation parameters. This rest state is a predefined pose configuration for each object instance~(\eg all drawers closed, microwave shut), set during the data construction. In contrast to parameterizing motion relative to the first observed frame—which is sequence-dependent and thus inconsistent—this canonicalization ensures identical ground truth across different sequences of the same object for both part bounding boxes and underlying geometry/texture. This results in more stable training and substantially faster convergence, an important benefit given the limited availability of articulated 3D data.

\subsection{Model Architecture}
\label{s:method-architecture}

Given the multi-view, multi-stage inputs described above, \model maps images into a shared token space, routes these token features to a fixed set of learnable part slots, and decodes for each part both the geometry/texture plane representations and articulation structure parameters. 

\myparagraph{Encoding image tokens.}
Each image $\bI_{v,t}$ is first divided into non-overlapping $8{\times}8$ patches and projected by a small MLP~\cite{popescu2009multilayer} into a sequence of image tokens. To disambiguate tokens across views and articulation stages, we augment them with three types of side information.

\vspace{0.1cm}
\noindent
\underline{Stage embeddings.} 
We add a learnable embedding $\be_t$ to every token originating from stage $t$, enabling the network to separate information from different articulation states.

\vspace{0.1cm}
\noindent
\underline{Viewpoint information.}
For each input image we compute the Pl\"{u}cker ray representation~\cite{jia2020plucker,sitzmann2021light} using known camera intrinsics and extrinsics, denoted as $(\mathbf{v},\mathbf{v}\times\mathbf{o})$, where $\mathbf{v}$ is the unit ray direction and $\mathbf{o}$ is the camera origin.

\vspace{0.1cm}
\noindent
\underline{High-level semantics.}
We concatenate features from a pretrained DINOv2 encoder~\cite{oquab2023dinov2}; the image inputs to DINOv2 are resized to account for different patch sizes. These features provide rich semantics~\cite{li2024learning,el2024probing,zhang2023tale,xu20254dgt},  which is especially helpful under sparse views and limited training data.

After the concatenation of the above features with patch embeddings, we obtain the final token sequence from the input images, serving as the conditional input to the transformer layers.

\begin{figure}
    \centering
    \includegraphics[width=\linewidth]{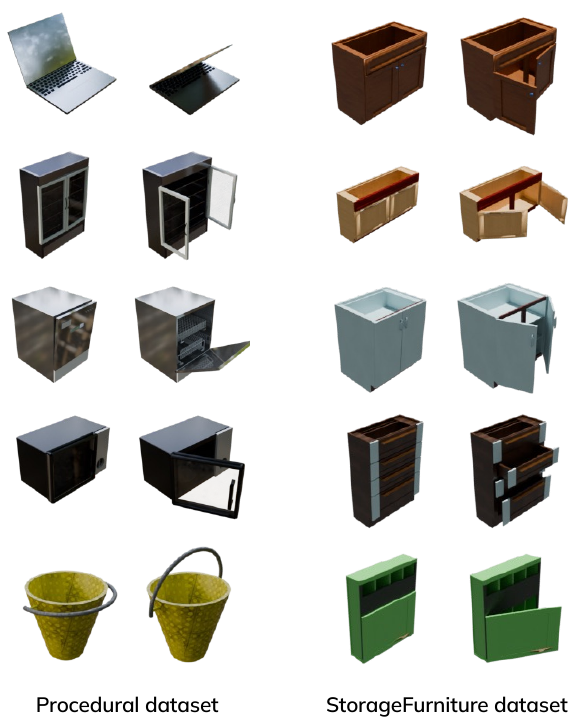}
    \caption{\textbf{Articulated object data samples} from two of our data collections: Procedural~(left two columns) and StorageFurniture~(right two columns). For each object, we show two states from one dynamic sequence under the same viewpoint.}
    \label{fig:dataset}
    \vspace{-2mm}
\end{figure}

\myparagraph{Learnable part slots.}
We introduce $P_0$ learnable part slots in the model. Each slot has a set of tokens trained to predict part information given the articulated object observation inputs. One slot is reserved for the base part, and the remaining slots model movable parts. During training, slot tokens are updated via attention with image tokens~\cite{vaswani2017attention}. With part number $P$ given, we simply keep the first $P$ slots to obtain the final prediction in both training and inference.

\myparagraph{Transformer layers.} 
Inputs to the transformer are part tokens and image tokens~(Figure~\ref{fig:method}). Stacked layers update part tokens while continuously progressing on information in image tokens.

We employ two complementary layer types. In the \emph{self-attention} layers, we concatenate image and part tokens and apply a single attention operation over the entire set. This promotes global context sharing across views, stages, and parts, which is beneficial for long-range correspondence and enforcing inter-part consistency. In the \emph{cross-attention} layers, image tokens act as queries and part tokens provide keys and values, explicitly routing visual information to a compact set of part slots and mitigating inter-part interference—an important factor to further improve performance.

Most LRM-style models~\cite{hong2023lrm,xu2024grm,wei2024meshlrm,li2025lirm} adopt only self-attention layer. In our setting this led to slower training and unstable specialization of part slots, so we replace ${\sim}75\%$ of layers with cross-attention, for two reasons:
(1) \emph{Token efficiency.} Multi-stage, multi-part inputs introduce many more tokens than single-object settings; cross-attention uses a smaller effective attention window and is more efficient.
(2) \emph{Convergence and accuracy.} Interleaving cross-attention accelerates convergence and improves final quality by encouraging distinct roles for image vs.\ part tokens, focusing the model on their interactions, and enabling it to learn stronger reconstruction priors.

\myparagraph{Decoding part properties.} 
As shown in Fig.~\ref{fig:method}, we split the final part tokens into two branches and use separate MLP heads to predict (i) the hexa-plane representation $\mathcal{T}_p$ for geometry/texture and (ii) the articulation vector $\hat{\mathcal{A}}_p\in\mathbb{R}^{14+T}$. We partition $\hat{\mathcal{A}}_p$ along the channel axis into
$(\hat{\bB}_p,\hat{\bC}_p,\hat{\bD}_p,\hat{\bO}_p,\hat{\bS}_p)$
and remap these raw outputs to the final properties:
\begin{align}
    \bB_p &= (2r \cdot \psi(\hat{\bB}_{p,\mathrm{center}})- r, 2r \cdot \psi(\hat{\bB}_{p,\mathrm{size}})) \\
    \bD_p &= \hat{\bD}_{p} / ||\hat{\bD}_{p}||_2 \\
    \bO_p &= 2r \cdot \psi(\hat{\bO}_p) - r \\
    \bS_p &= 2 \cdot \psi(\hat{\bS}_p) - 1,
\end{align}
where $\psi(\cdot)$ is the sigmoid and $r$ is the radius of the normalized bounding sphere. For the motion type, $\hat{\mathbf{C}}_p$ produces two logits (i.e., prismatic vs.\ revolute), which is determined by \texttt{softmax} during training and \texttt{argmax} at inference. By convention, the first part slot is reserved for the static base with motion type fixed to $\mathrm{static}$. The two-way classification is only applied to the remaining $P_0{-}1$ movable slots.

\subsection{Rendering Articulated Object}
\label{s:method-rendering}
As shown in Fig.~\ref{fig:method}, during each training iteration we render the articulated object both per part and as a composite of all parts to generate the final supervision image. Our renderer follows signed-distance-function (SDF) volume rendering~\cite{yariv2021volume} to allow the model to learn both underlying geometry and appearance. Specifically for each dynamic part, we transform sampling rays into the object coordinate space of the corresponding state. Implementation details for rendering static and dynamic parts and for composing \emph{all} parts via volumetric rendering are provided in Sec~\ref{supp:method-rendering} in the supplementary.

\subsection{Training Scheme}
\label{s:training-scheme}
\myparagraph{Training objectives.}
Our loss combines \emph{rendering} objectives with \emph{direct} supervision on articulation parameters. --- All rendering losses are computed on \emph{per-part} renderings rather than the final composite. Empirically, supervising only the composite image hinders learning in occluded regions and biases geometry/texture near part boundaries. For each part $p$, view $v$, and stage $t$, we apply mean-squared error $\mathcal{L}_2$ on RGB and masks, and a perceptual loss $\mathcal{L}_{\text{LPIPS}}$~\cite{zhang2018unreasonable} on RGB. For articulation parameters, we use cross-entropy $\mathcal{L}_{\text{CE}}$ for motion-type classification $C_p$ and MSE for the remaining parameters $B_p, D_p, O_p, S_p$.

\myparagraph{Pre-training stage.}
Articulated-object datasets are inherently less diverse and scalable than static 3D corpora~\cite{deitke2023objaverse,deitke2023objaversexl}. To learn a strong prior over geometry, texture, and part decomposition, we introduce an optional pre-training stage. 
For pre-training, we curate $130$k static 3D objects with part decomposition from a collection of 3D-artist generated assets that we licensed for AI training from a commercial source. And we further
filter the assets whose native glTF/GLB hierarchy contains at most $P_0$ parts (as defined by their mesh-based composition). During pre-training, the model is optimized only with rendering losses ($\mathcal{L}_2$, $\mathcal{L}_{\text{LPIPS}}$) and MSE on part bounding-box centers/sizes; articulation parameters (motion type, axis, dynamics) are not applicable. Empirically, this stage consistently boosts downstream performance across all metrics.

\myparagraph{Coarse-to-fine articulation training stage.}
Following static pre-training, we fine-tune \model on our articulated dataset using a coarse-to-fine curriculum designed to gradually increase the network ability. First, we gradually sharpen the rendered surfaces by linearly increasing the reciprocal of the standard derivation of the SDF~\cite{li2025lirm}. Second, we employ resolution annealing: training begins at a $128{\times}128$ resolution for rendering-based objectives, and the supervision resolution is later increased to $256{\times}256$ to encourage finer geometric predictions.

\subsection{Articulated Object Datasets}
Our model is trained with multi-view, multi-state RGB images from dynamic articulated sequences generated from a large and diverse collection of articulated 3D object assets. Complete details of the data pipeline (asset collection/construction and per-asset sequence generation) and statistics are given in the supplementary material. Below we briefly introduce the names and characteristics of each collection used. In total, we aggregate articulated objects from three primary sources:

\myparagraph{PartNet-mobility.}
We utilize several common indoor categories from the PartNet-Mobility benchmark~\cite{xiang2020sapien}, including buckets, microwaves, and a range of furniture classes.

\myparagraph{Procedural dataset.}
To enhance diversity and realism, we generated a new articulated-object dataset using a procedural generation method motivated by ~\cite{joshi2025proceduralgenerationarticulatedsimulationready}. It includes $2{,}000$ high-quality articulated models across six categories, with rich variation in shape and texture.

\myparagraph{StorageFurniture dataset.}
We specifically focus on this category because it is ubiquitous in real-world environments and widely used as a training domain in prior work~\cite{chen2024urdformer,liu2024singapo,jiang2022opd}. Our collection is procedurally generated from the PartNet-Mobility storage-furniture category: using a compositional assembly system, we replace parts of source objects with alternative geometries from other assets, yielding a large and diverse set of realistic storage-furniture models.

Figure~\ref{fig:dataset} shows random samples from the latter two collections. For every articulated asset described above, we automatically synthesize multiple articulated sequences, which are used for training and evaluation.

\section{Experiments}%
\label{s:experiments}

\myparagraph{Evaluation datasets.}
We evaluate on two distinct test sets.

\noindent\textbf{StorageFurniture Test Set:} For comparisons to feed-forward methods, we use the held-out split of our large-scale StorageFurniture dataset, comprising $631$ objects, enabling robust large-scale evaluation.

\noindent\textbf{PartNet-mobility Test Set:} For comparisons to optimization-based methods, we use a held-out set of $10$ articulated sequences from PartNet-mobility~\cite{xiang2020sapien}, aligning with the evaluation protocol of previous optimization methods on small sets due to their long inference times.

\myparagraph{Baseline methods.}
We compare against two classes of methods.
For feed-forward models, we select URDFormer~\cite{chen2024urdformer} and SINGAPO~\cite{liu2024singapo}, two recent state-of-the-art methods for articulated reconstruction with a particular focus on part-level articulation prediction for storage-furniture objects. For optimization-based methods, we include PARIS~\cite{liu2023paris}, DTA~\cite{weng2024neural} and ArtGS~\cite{liu2025building}, which recover geometry, texture, and articulation structure from two-state, multi-view inputs.

\myparagraph{Evaluation metrics.}
To provide a holistic evaluation, we assess geometry, texture, and part-level accuracy.
\begin{itemize}
    \item Image-level metrics: We report PSNR and LPIPS~\cite{zhang2018unreasonable} on renderings from $8$ novel test viewpoints.
    \item Geometry-level metrics: We compute Chamfer Distance~(CD) and F-Score between the predicted and ground-truth meshes.
    \item Part-level metrics: Following the evaluation protocol in \citet{liu2024singapo}, we match predicted parts and ground-truth parts via Hungarian algorithm, then report (i) part-level prediction accuracy using distance of generalized Intersection over Union~(1-gIoU)~\cite{rezatofighi2019generalized} and (ii) Euclidean distance between part centroids. For both, lower is better.
\end{itemize}

\myparagraph{Implementation details.}
We train two versions of \model: a multi-view model~($V{=}4$) for comparison with optimization methods and a monocular model~($V{=}1$) for fair comparison with feed-forward baselines. We fix $T{=}2$~(``start \& end") states. All training and inference images are resized to $H{=}W{=}128$ to balance compute.

\subsection{Results}

\begin{figure*}
    \centering
    \includegraphics[width=\linewidth]{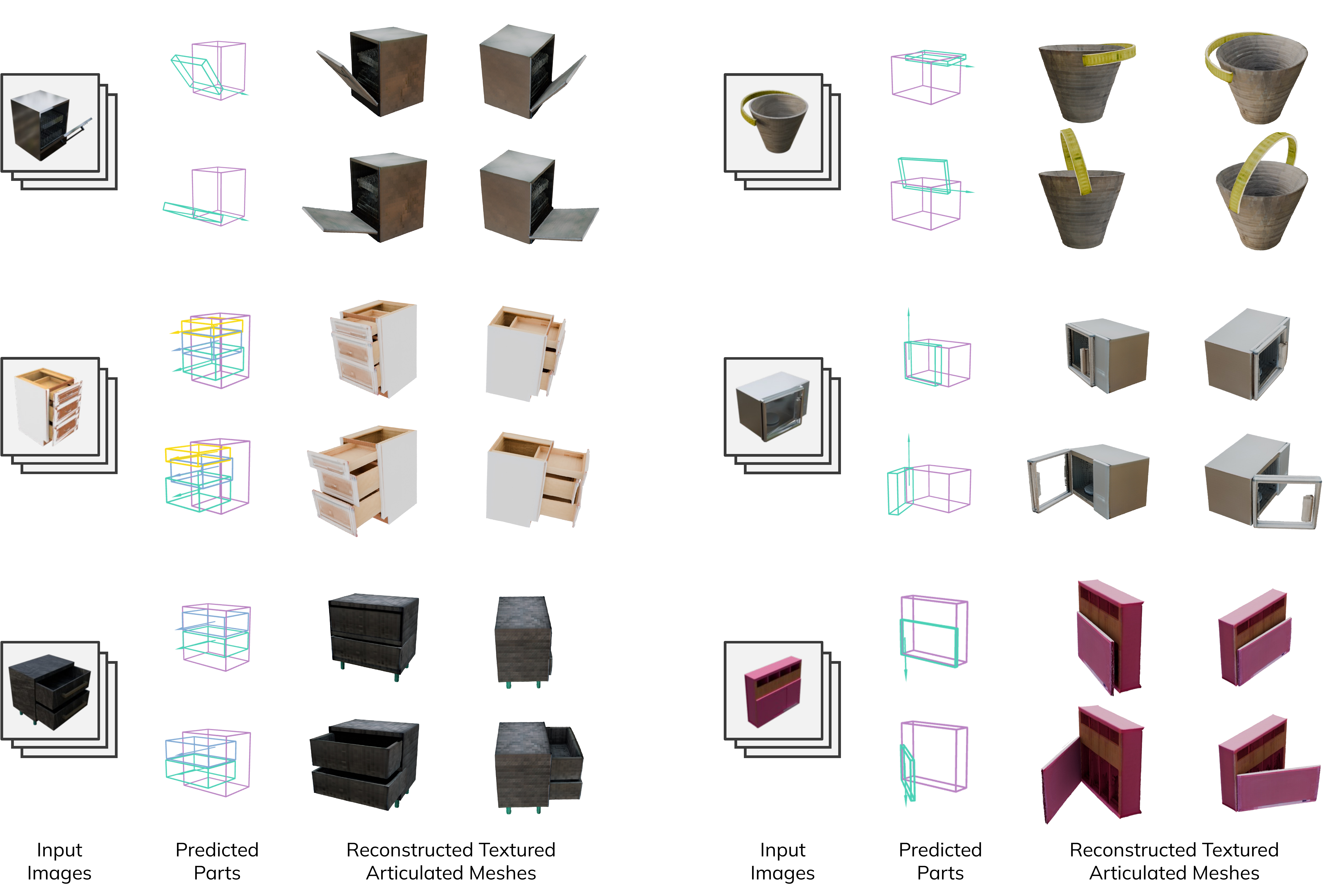}
    \caption{\textbf{Qualitative results.} ``Input Images" shows the multi-state inputs provided to the model~(visualized by one image). ``Predicted Parts" visualizes the predicted per-part bounding boxes in the start and end states; ``Reconstructed Textured Articulated Meshes" displays the textured 3D reconstructions for the same two articulated states.}
    \label{fig:qualitative}
\end{figure*}

\begin{table}[t]
    \small
	\setlength{\tabcolsep}{12pt} %
	\centering
	\begin{tabular}{lccc}
    \toprule
        & $d_{\mathrm{gIoU}}$~$\downarrow$ & $d_{\mathrm{cDist}}$~$\downarrow$ & CD~$\downarrow$   \\
    \midrule
    URDFormer~\cite{chen2024urdformer}
    & 1.0710 & 0.1622 & 0.0536 \\
    SINGAPO~\cite{liu2024singapo} 
    & \underline{0.8306} & \underline{0.0947} & \underline{0.0059} \\
    \model~(Ours) &  
    \textbf{0.4717} &  
    \textbf{0.0538} &  
    \textbf{0.0019} \\
    \bottomrule
    \end{tabular}
	\caption{\textbf{Quantitative comparison} on the StorageFurniture test set (all metrics: lower is better). Our method surpasses baselines in both part-level prediction and holistic geometry reconstruction.}
\label{tab:sf-feedforward}
\end{table}

\begin{table}[t]
    \small
	\setlength{\tabcolsep}{8pt} %
	\centering
	\begin{tabular}{lcccc}
    \toprule
        & PSNR~$\uparrow$ & 
        LPIPS~$\downarrow$ & 
        CD~$\downarrow$ & 
        F-Score~$\uparrow$ \\
    \midrule
    PARIS~\cite{liu2023paris} &
    \underline{22.851} &
    0.183 &
    0.023 &
    0.486 \\
    DTA$^*$~\cite{weng2024neural} &
    21.587 &
    \underline{0.165} &
    \textbf{0.008} &
    \textbf{0.821} \\
    ArtGS~\cite{liu2025building} &
    22.352 &
    0.176 &
    0.016 &
    0.520 \\
    \model~(Ours) &
    \textbf{27.059} &
    \textbf{0.049} &
    \underline{0.009} &
    \underline{0.762} \\
    \bottomrule
    \end{tabular}
	\caption{\textbf{Quantitative comparison} of the reconstruction quality on the PartNet-mobility~\cite{xiang2020sapien} test set. $^*$~indicates requiring additional depth inputs.}
\label{tab:pm-optimzation}
\end{table}

\begin{figure}[t]
  \centering
   \includegraphics[width=\linewidth]{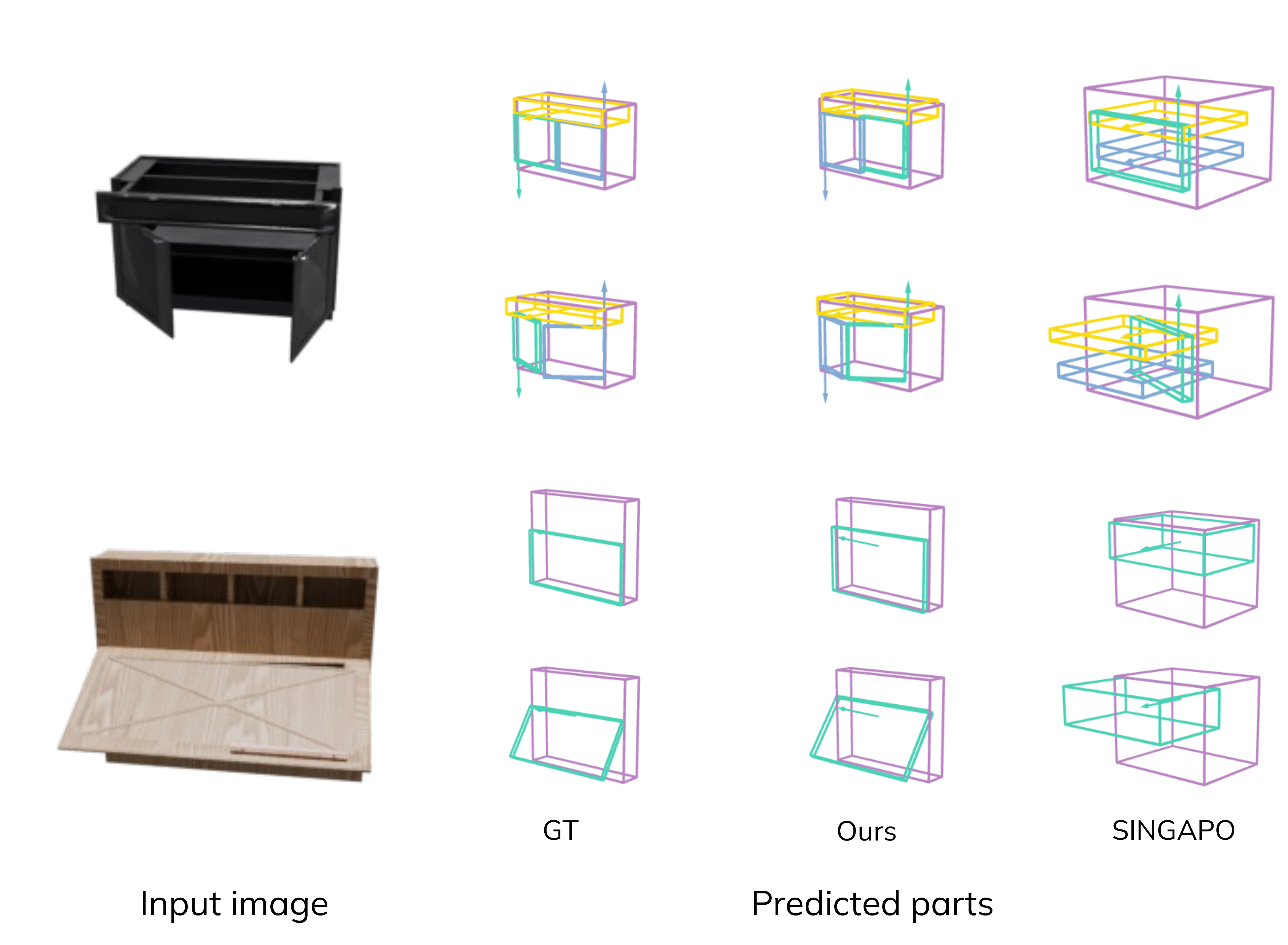}

   \caption{\textbf{Qualitative comparison with feed-forward baseline} of predicted part bounding box and articulation structure.}
   \label{fig:comparison-ff}
   \vspace{-2mm}
\end{figure}
\begin{figure}[t]
  \centering
   \includegraphics[width=\linewidth]{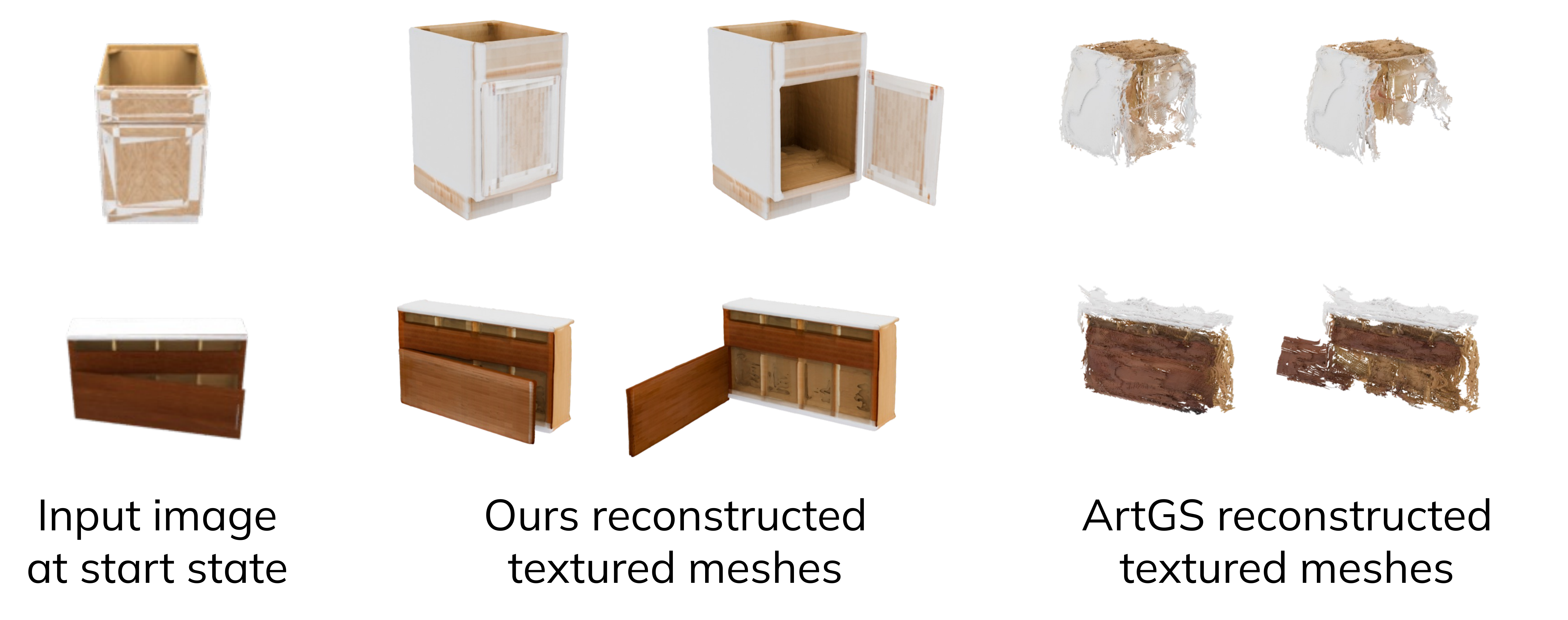}

   \caption{\textbf{Qualitative comparison with optimization baseline.} With the same sparse inputs, our method reconstructs complete, high-fidelity textured meshes in both states. ArtGS~\cite{liu2025building} lacks reliable correspondences under sparsity, leading to fragmented and inaccurate results.}
   \label{fig:comparison-optim}
\end{figure}

\myparagraph{Comparison with feed-forward baselines.}
We first compare the monocular version of \model to  URDFormer and SINGAPO. Table~\ref{tab:sf-feedforward} reports quantitative results: \model outperforms both baselines by a large margin across all metrics. In particular, it achieves substantially better part-level prediction, as evidenced by lower $d_{\mathrm{gIoU}}$ and $d_{\mathrm{cDist}}$—and more accurate overall geometry, as indicated by a reduced Chamfer Distance. These gains are also evident qualitatively in Figure~\ref{fig:comparison-ff}, which contrasts our predicted part bounding boxes with SINGAPO and highlights \model’s enhanced ability to jointly reconstruct part structure and geometry.

\myparagraph{Comparison with optimization-based baselines.}
Next, we evaluate the multi-view \model against per-object optimization baselines on the PartNet-mobility test set, using sparse inputs (4 views across 2 states) for all methods; DTA additionally receives depth maps, as required. As shown in Table~\ref{tab:pm-optimzation}, \model achieves state-of-the-art results on image-level metrics, outperforming all optimization-based competitors by a clear margin. While DTA reports comparable geometry metrics, this is expected given its depth supervision; nonetheless, its lower PSNR and higher LPIPS scores indicate poor appearance recovery. This disparity is rooted in sparsity: optimization approaches rely on dense cross-state correspondences, which are fragile and difficult to establish from few views. As shown by the failure example in Figure~\ref{fig:comparison-optim}, due to the lack of robust correspondences, ArtGS yields fragmented, noisy geometry, whereas \model—benefiting from a strong learned prior—reconstructs coherent, high-fidelity textured meshes.

\myparagraph{Qualitative results.}
We present qualitative results in Figure~\ref{fig:teaser} and Figure~\ref{fig:qualitative}. Our core idea is to cast articulated object reconstruction as a part-based prediction problem, decomposing objects into a set of rigid parts. As shown in Figure~\ref{fig:teaser}, \model successfully reconstructs diverse object categories and yields clear part decompositions, visualized by rendering each predicted part with a consistent color.

Figure~\ref{fig:qualitative} provides a closer look at our method's outputs, showing reconstructions for the start/end states provided in the image inputs. The \emph{Predicted Parts} column visualizes the inferred structure by overlaying per-part bounding boxes for both states, along with recovered articulation structures (e.g., axes for movable parts). Together, these visualizations demonstrate that \model reconstructs high-fidelity geometry and texture while accurately recovering the articulation structure that drives object motion.

\begin{figure}[t]
  \centering
   \includegraphics[width=\linewidth]{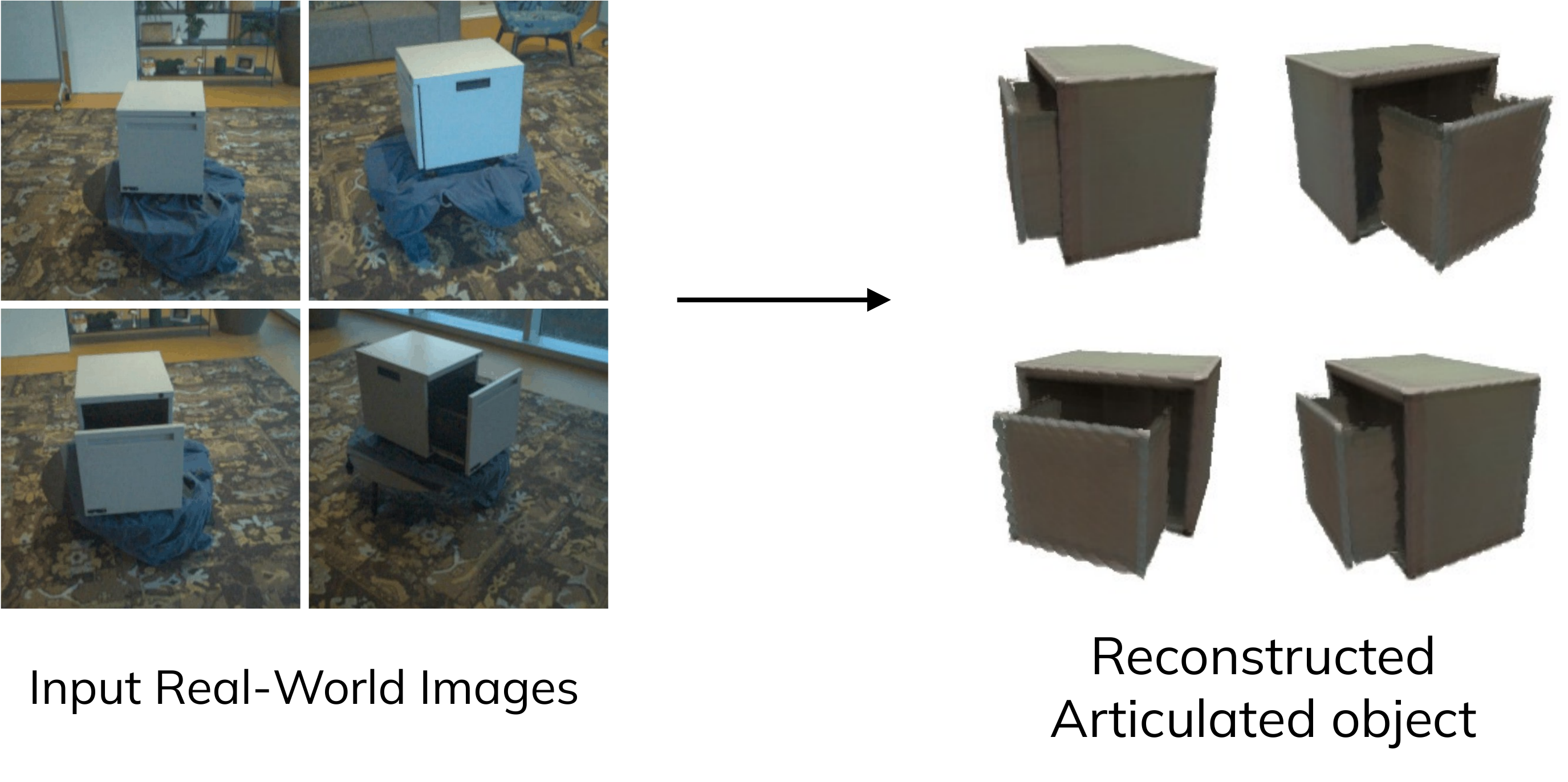}
   \caption{\textbf{Real-world images results.} }
   \label{fig:real-world}
   \vspace{-2mm}
\end{figure}

\myparagraph{Real-world images results.}
We further present a real-world example in Fig.~\ref{fig:real-world}. The capture uses only approximate camera poses and no background masking. Despite no real-image training, \model recovers the correct articulation structure with plausible geometry and texture.

\subsection{Ablation Study}
\begin{table}[t]
    \footnotesize
	\setlength{\tabcolsep}{1pt} %
	\centering
	\begin{tabular}{lcccccc}
    \toprule
    & PSNR~$\uparrow$ & 
    LPIPS~$\downarrow$ & 
    CD~$\downarrow$ & 
    F-Score~$\uparrow$ &
    $d_{\mathrm{gIoU}}$~$\downarrow$ & $d_{\mathrm{cDist}}$~$\downarrow$  \\
    \midrule
    monocular view &
    25.961&
    0.037&
    0.017&
    0.591&
    0.731&
    0.086\\
    w/o rest-state &
    23.587&
    0.081&
    0.015&
    \underline{0.672}&
    0.878&
    0.113\\
    
    w/o defined part order &
    22.588&
    0.060&
    0.015&
    0.564&
    1.118&
    0.208\\

    w/o part rendering loss &
    24.465&
    0.047&
    0.022&
    0.512&
    0.681&
    \underline{0.075}\\
    \midrule
    w/o pre-train &
    \underline{27.495}&
    \underline{0.034}&
    \underline{0.013}&
    0.641&
    \underline{0.665}&
    0.082\\
    \midrule
    \model~(Ours) &
    \textbf{28.678}&
    \textbf{0.030}&
    \textbf{0.008}&
    \textbf{0.716}&
    \textbf{0.629}&
    \textbf{0.062}\\
    \bottomrule
    \end{tabular}
	\caption{\textbf{Ablation study} of the components in proposed \model.}
\label{tab:ablation}
\end{table}
We conduct a detailed ablation study (Table~\ref{tab:ablation}) to validate our key design choices.All model variants are evaluated on the overall held-out test set of PartNet-mobility~\cite{xiang2020sapien}, which includes more than $130$ multi-view, multi-state sequences. Our reference is the \textit{base model w/o pre-train}, which is the multi-view~($V{=}4$) \model model trained from scratch on only the articulated object dataset. We then measure the impact of adding the pre-training stage (last row) and of removing key components (first four rows) relative to this base model.

\myparagraph{Pre-training stage helps.}
By adding the pre-training stage, the full \model shows a substantial improvement over the version without pre-training, with gains across geometry, texture, and articulation metrics. This confirms that a robust prior on geometry and part decomposition learned from large-scale 3D data is highly beneficial for part-based articulated reconstruction.

\myparagraph{Monocular \vs multi-view inputs.}
We compare the base multi-view model with a variant trained using only a single view as input. This change produces a clear drop in quality across all metrics, as expected, confirming that multiple views are crucial for resolving ambiguities and achieving high-fidelity, geometrically accurate reconstruction.

\myparagraph{Rest-state formulation.}
Removing the rest-state formulation and predicting parts relative to the first observed frame leads to a severe quality drop~(PSNR $27.495\to23.587$) and higher part-level errors ($d_{\mathrm{cDist}}$ $0.082\to0.113$). The decline is unsurprising: because the test set contains multiple sequences of the same object, the first-frame scheme induces a pose-dependent “canonical” per sequence (e.g., one starting closed, another open), effectively treating the same object as different identities. This ambiguity hampers learning and underscores the benefit of a consistent, predefined rest state.

\myparagraph{Defined Part Order.}
Without forcing a pre-defined part ordering in dataset construction, performance degrades most severely across the board~(PSNR $27.495\to22.588$, $d_{\text{cDist}}$ $0.082\to0.208$). This is expected: Lacking a consistent ordering forces the network to learn both reconstruction and part matching, which is especially difficult for objects with many similar dynamic parts (e.g., cabinets with many drawers). We observe slot collapse—multiple slots predicting parts in the same location—directly reflected in the part-level metrics. This confirms the benefit of enforcing a consistent part ordering.

\myparagraph{Per-part \vs composite rendering losses.}
Finally, eliminating per-part rendering losses and supervising only the composited image consistently harms performance. Per-part supervision is essential for learning occluded content; otherwise, the result is smeared geometry and inferior texture quality.

\section{Discussions and Conclusion}
\label{s:conclusion}
\myparagraph{Limitations.}
Our method assumes a known part count for the target object and relies on pre-calibrated camera poses. Future work should include learning a pose-free variant (self-calibrated cameras) with larger datasets and integrating part-count estimation directly into the model.

\myparagraph{Conclusion.}
In this work, we propose \model, a feed-forward model to reconstruct complete 3D articulated objects from sparse, multi-state images. By casting reconstruction as a part-based prediction problem, ART jointly decodes geometry, texture, and articulation structure for each part. Experiments across a broad range of articulated objects demonstrate that \model consistently outperforms strong feed-forward and optimization-based baselines.

\myparagraph{Acknowledgments.}
We thank Yawar Siddiqui, Ruocheng Wang, and Qiao Gu for the comments and fruitful discussions, Ka Chen, David Clabaugh, and Samir Aroudj for the help on dataset collecting and constructions, and Yanjie Ze and Pengyu Mo for the robot simulation results. This work is in part supported by NSF RI \#2338203 and ONR MURI N00014-22-1-2740.

{
    \small
    \bibliographystyle{ieeenat_fullname}
    \bibliography{main}
}

\clearpage
\renewcommand\thefigure{S\arabic{figure}}
\setcounter{figure}{0}
\renewcommand\thetable{S\arabic{table}}
\setcounter{table}{0}
\renewcommand\theequation{S\arabic{equation}}
\setcounter{equation}{0}
\pagenumbering{arabic}%
\renewcommand*{\thepage}{S\arabic{page}}
\setcounter{footnote}{0}
\setcounter{page}{1}
\maketitlesupplementary
\appendix

\section{More results}

We provide additional results in the project \href{https://kyleleey.github.io/ART/}{webpage}.

\myparagraph{Video results.}
We provide both fixed-view and rotational-view renderings of the reconstructed articulated objects, with the dynamic parts moving according to the predicted articulation structure.

\myparagraph{Export into simulator.}
As discussed in the main text, our part-based output can be directly converted to URDF. Combining the exported URDF with each part’s textured mesh yields simulation-ready assets. We showcase several interaction scenes in the MuJoCo simulator—each featuring a humanoid robot and an articulated object—and include the corresponding videos in the project \href{https://kyleleey.github.io/ART/}{webpage}.

\myparagraph{Detailed runtime analysis.}
We provide a quantitative runtime comparison on a single A100 GPU in Table~\ref{tab:rebuttal-runtime}.
ART completes reconstruction in $0.85$s~(multi-view) and $0.42$s~(monocular). This is orders of magnitude faster than optimization baselines with lengthy per-instance processing. This efficiency stems in part from ART's use of cross-attention layers, which reduces token processing time.

\begin{figure}[t]
  \centering
   \includegraphics[width=\linewidth]{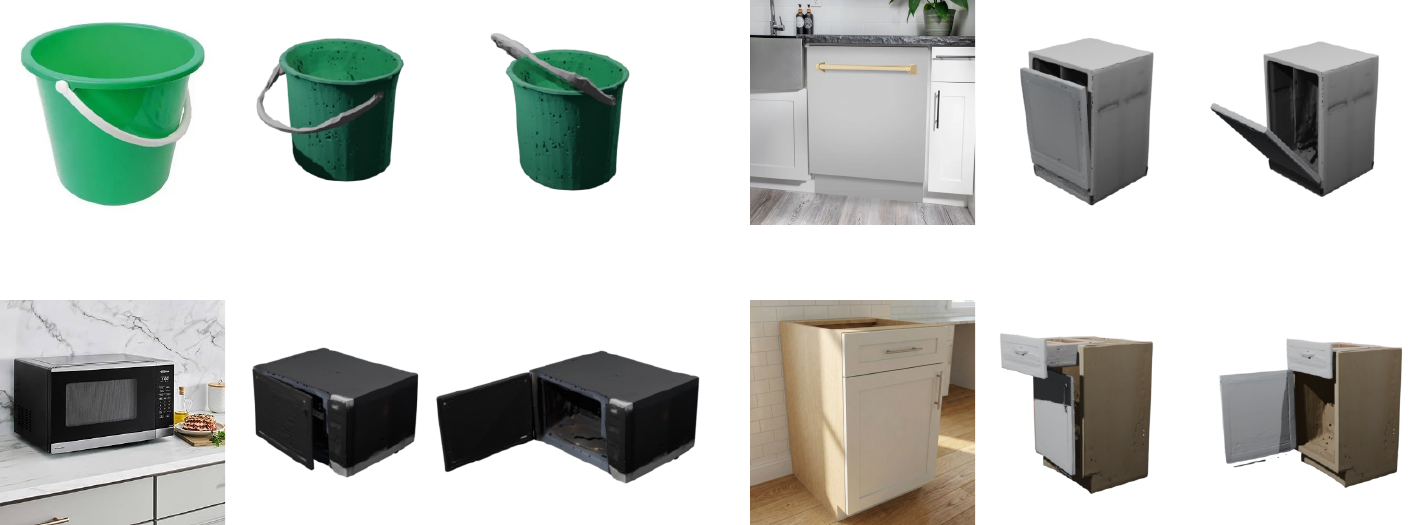}
   \caption{\textbf{Real-world images results.} }
   \label{fig:rebuttal-real-world}
\end{figure}

\myparagraph{Real-world results}
Here we add four additional real-world results~(Figure~\ref{fig:rebuttal-real-world}). While texture reconstruction is inherently limited by the synthetic training domain, ART demonstrates robust articulation-structure understanding across diverse opening configurations.

\section{Rendering Articulated Object}
\label{supp:method-rendering}

At each training iteration, we render the articulated object both per part and in a compositional manner to obtain the final image for supervision. The rendering process queries hexa-plane feature outputs from the model to compute the per-pixel RGB and uses the predicted articulation to correctly place each dynamic part.

Our rendering pipeline follows the signed distance function~(SDF) volume rendering~\cite{yariv2021volume}. For a camera ray defined by origin $\mathbf{o}$ and unit direction $\mathbf{v}$, we first intersect the ray with the part’s axis-aligned bounding box $\bB_p$~(in canonical space) and sample 3D points along the valid ray segment.  Each sampled world-space point $\mathbf{x}_p(\mathbf{o},\mathbf{v},\mathbf{B}_p)$ is mapped to the part’s normalized local coordinates 
$\hat{\mathbf{x}}_p=(x,y,z)\in[-1,1]^3$, which are then used to query the hexa-plane features:
\begin{equation}
    \mathbf{f}_{p,xy} = 
\begin{cases}
\mathrm{Bilinear}(\mathbf{T}_{pxy+}; x, y), & z \ge 0,\\[4pt]
\mathrm{Bilinear}(\mathbf{T}_{pxy-}; x, y), & z < 0.
\end{cases}
\end{equation}

Here $\mathbf{T}_{pxy+}$, $ \mathbf{T}_{pxy-}$ are the model's hexa-plane representation output from Eq.~\ref{eq:hexa-plane}; $\mathbf{f}_{p,yz}$, $\mathbf{f}_{p,xz}$ can be obtained analogously. These features are concatenated to form the feature vector $\mathbf{f}_p$ for the spatial point $\mathbf{x}+p$. Following the architecture in \cite{wei2024meshlrm,li2025lirm}, we use two small MLPs to predict SDF value $\mathbf{s}_p$ and RGB color $\mathbf{c}_p$, respectively. In particular, the SDF value is computed as:
\begin{equation}
   \mathbf{s}_p = \mathrm{MLP}(\mathbf{f}_p) + \mathbf{s}_{\mathrm{bias}}(\mathbf{x}_p).
\end{equation}
where the bias term $\mathbf{s}_{\mathrm{bias}}(\mathbf{x}_p) = ||\mathbf{x}_p|| - 0.1r$ is a prior defined in the part's local space, to initialize the shape as a sphere to stabilize the training~\cite{yariv2021volume,li2025lirm}. The SDF value $\mathbf{s}_p$ can be converted further to volume density using the Laplace CDF~\cite{yariv2021volume}:
\begin{equation}
    \sigma_p = 
    \begin{cases}
        \frac{1}{2}\mathrm{exp}(-\frac{\mathbf{s}_p}{\beta}), & \mathbf{s}_p \ge 0,\\[4pt]
        1 - \frac{1}{2}\mathrm{exp}(-\frac{\mathbf{s}_p}{\beta}), & \mathbf{s}_p < 0.
    \end{cases}
    \label{eq:sdf-density}
\end{equation}

 where $\beta$ is the standard deviation that controls the sharpness of the underlying surface. As noted in the main text, $\frac{1}{\beta}$ is linearly annealed over training to progressively sharpen surfaces. With $\sigma_p$ and $\mathbf{c}_p$ evaluated at ray samples, we compute \emph{per-part} color, opacity (mask), and depth via the usual transmittance accumulation~\cite{mildenhall2021nerf,yariv2021volume}. Normals are obtained by numerically differentiating the SDF at each sample~\cite{li2023neuralangelo}.

To obtain the composited image over \emph{all parts}, we merge the sampled points from all parts along each ray, sort their orderings by the sampled ray distance, and apply standard alpha compositing to compute the rendered images. We employ Nerfacc~\cite{li2022nerfacc} to accelerate point sampling, ray distance sorting, and compositing.

\begin{table}[t]
    \footnotesize
	\centering
	\begin{tabular}{lccc}
    \toprule
    Optimization-based & PARIS & 
    DTA & 
    ArtGS \\
    Runtime~(min) &
    $50$&
    $40$&
    $9$ \\
    \midrule
    Feed-forward &
    URDFormer &
    SINGAPO & Ours  \\
    Runtime~(s)
    &
    $2.37$ &
    $0.54$ &
    $0.85$ / $0.42$\\
    
    \bottomrule
    \end{tabular}
	\caption{\textbf{Runtime comparison} between baselines and \model.}
\label{tab:rebuttal-runtime}
\end{table}

\myparagraph{Dynamic parts rendering.}
The equations above describe static per-part rendering in each part's canonical space~(from the model-predicted bounding-box region). For dynamic parts at stage $t$, instead of physically moving the part’s volume and rebuilding an oriented bounding box, we transform the camera rays into the part’s instantaneous local frame (\ie move the rays inversely), keeping the axis-aligned bounding box $\mathbf{B}_p$ unchanged.

Recall that $\mathbf{C}_p$ denotes the motion type, $\mathbf{D}_p$ the unit joint axis, $\mathbf{O}_p$ a point on that axis, and $\mathbf{S}_{p,t}\in[-1,1]$ the normalized motion value at stage $t$. For a point $\mathbf{x}$ in the part’s canonical space, the stage-$t$ rigid transform $T_{p,t}$ for different motion types can be defined as:

\textbf{Prismatic:}
\begin{equation}
T_{p,t}(\mathbf{x}) \;=\; \mathbf{x} \;+\; (2r)\,\mathbf{S}_{p,t}\,\mathbf{D}_p .
\end{equation}

\textbf{Revolute:}
\begin{equation}
T_{p,t}(\mathbf{x}) \;=\; \mathbf{O}_p \;+\; R\!\big(\mathbf{D}_p,\; 2\pi\,\mathbf{S}_{p,t}\big)\,\big(\mathbf{x}-\mathbf{O}_p\big),
\end{equation}
where $R(\mathbf{d},\theta)$ denotes the rotation by angle $\theta$ (radians) about axis $\mathbf{d}$.

Now, instead of actually transforming the bounding box~(point), we'll inversely transform each ray $(\mathbf{o}, \mathbf{v})$ to reach the equivalent rendering results. It's straightforward that for prismatic parts, after transformation:
\begin{equation}
    \hat{\mathbf{o}} = \mathbf{o} - 2r \cdot \bS_{p,t} \cdot \bD_p, \quad \hat{\mathbf{v}} = \mathbf{v},
\end{equation}
and for revolute parts, similarly:
\begin{align}
    \hat{\mathbf{o}} &= \mathbf{o} + \mathrm{Rotate}(\bD_p, -2\pi \cdot \bS_{p,t})(\mathbf{o} - \bO_p) + \bO_p,\\ 
    \hat{\mathbf{v}} &= \mathrm{Rotate}(\bD_p, -2\pi \cdot \bS_{p,t})\mathbf{v}.
\end{align}
We then proceed exactly as in the static case: intersect the transformed ray $(\hat{\mathbf{o}}, \hat{\mathbf{v}})$ with bounding box $\bB_p$, sample points, query hexa-planes, and composite to obtain the desired renderings. The computed values from the queried features are further assigned to the original ray sampled points, which are in the current world coordinate space, for volume rendering composition.

\section{Dataset Construction}

A key advantage of \model is its training data: we substantially increase both the quantity and diversity of articulated assets. Prior feed-forward models~\cite{liu2024singapo,jiang2022opd} typically rely on subsets of PartNet-Mobility~\cite{xiang2020sapien}, which offer limited geometric diversity and often unrealistic textures.

In this work, we combine three articulated-object data sources to construct our training dataset, increasing diversity and fidelity in geometry, texture, and articulation complexity. Beyond the basic information in the main text, further details on these sources are provided below.

\myparagraph{PartNet-mobility.} 
As previously mentioned, PartNet-Mobility provides common indoor articulated categories—bucket, dishwasher, door, laptop, microwave, oven, refrigerator, and storage furniture. In total, we collect over $300$ objects from this dataset. However, many assets have unrealistic textures and low-quality surface geometry, motivating our exploration of more diverse and realistic articulated object sources.

\myparagraph{Procedural dataset.}
To scale both quantity and diversity, and inspired by Infinigen-Sim~\cite{joshi2025proceduralgenerationarticulatedsimulationready}, we adopt procedural generation to author articulated assets in Blender (with TA support). The resulting dataset includes $2{,}000$ high-quality articulated models across six categories: laptop, dishwasher, beverage refrigerator, cabinet with drawers, bucket, and microwave oven. 3D geometry is provided in GLB/OBJ and articulation in URDF. Each category is governed by procedural rules over shape, appearance, and articulation; for each, we generate several hundred variants with randomized shapes, sizes, and materials. In principle, this pipeline can produce an unlimited number of articulated objects.

\myparagraph{StorageFurniture dataset.} 
We recognize that the Storage-Furniture category in PartNet-Mobility spans many commonly used articulated assets, a property also noted by prior methods~\cite{chen2024urdformer,liu2024singapo} that primarily train on this category. Based on this, we construct a \textit{StorageFurniture} dataset by recombining parts from the PartNet-Mobility storage-furniture class.

For a given object, we use its articulation tree (kinematic structure) to procedurally create new instances via compositional part assembly: original parts are replaced with geometries from other objects (rescaled as needed), followed by UV-map correction and material randomization. This yields a large, realistic, and varied set of assets—over $10{,}000$ articulated models for training.

These three data sources collectively provide a large pool of articulated objects. Using each asset’s articulation definition, we further generate random per-part trajectories, resulting in a large dataset of articulated object sequences.

\section{More Implementation Details}

During sequence data construction, we explicitly set the order of the parts following a certain rule: the static base part is the first, and the remaining dynamic parts are ordered from low to high, front to back and left to right. This consistent ground-truth ordering greatly improves training stability and convergence speed. We also define the \emph{rest state} at this stage as the configuration where all dynamic parts are “closed.”

The transformer in \model has $32$ attention blocks with a $3\!:\!1$ cross-/self-attention ratio, $16$ heads, and a $256$-dimensional embedding. It outputs $24{\times}24$ hexa-plane features that are later upsampled to the spatial resolution of $192{\times}192$. We set the maximum part count to $P_0{=}8$. Training uses AdamW with $(\beta_1,\beta_2){=}(0.9,0.95)$. The multi-view model trains for $5$ days on $64$ H100 GPUs; the monocular model trains for $3$ days.

\model is trained with known camera poses as part of the input to distinguish tokens across views. This design follows common LRM-style practice, where camera information serves as implicit geometric cues for multi-view 3D reconstruction. Empirically, providing calibrated poses significantly accelerates training convergence and improves accuracy compared to using fully optimizable view embeddings. While we currently assume calibrated poses, we observe that with results remain reasonable under noisy real-world pose estimates. As discussed in our limitations, developing a pose-free variant by integrating optimizable view embeddings is a meaningful direction for future work to make the system more practical for arbitrary user-captured images.

\myparagraph{Inference details}
Similar to other LRM-style models, \model targets an object-centric reconstruction scenario. Consequently, it requires input images containing a single articulated object accompanied by a corresponding segmentation mask produced by off-the-shelf segmentation models. To prepare the input, we identify the longer side of the object's bounding box and crop a square region that preserves the full foreground, which is then resized to the input resolution. These standard pre-processing steps are used in both training and inference.

\end{document}